\documentclass[
	a4paper, 
	10pt, 
	unnumberedsections, 
	twoside, 
]{LTJournalArticle}

\runninghead{Supervised Machine Learning Models to Classify Aviation Safety Occurrences} 

\footertext{\textit{}} 

\usepackage{siunitx}
\usepackage{booktabs}
\usepackage[T1]{fontenc}
\usepackage{makecell}
\usepackage{adjustbox}
\usepackage{tabularx,ragged2e}
\usepackage{graphicx}
\usepackage{amsmath}

\title{A Practical Approach to using Supervised Machine Learning Models to Classify Aviation Safety Occurrences}

\author{
	Bryan Y. Siow\textsuperscript{1}\thanks{Corresponding author: \href{mailto:reccepython@gmail.com}{reccepython@gmail.com}\\
    \href{https://orcid.org/0009-0007-3278-9666}{https://orcid.org/0009-0007-3278-9666}}
}

\date{\footnotesize\textsuperscript{\textbf{1}}Independent Researcher, Singapore\\}

\renewcommand{\maketitlehookd}{
	\begin{abstract}
		\noindent This paper describes a practical approach of using supervised machine learning (ML) models to assist aviation safety investigators in classifying real-world aviation occurrences based on the factual details gathered from the event. Our implementation currently deployed as a ML web application identifies patterns within inputted occurrence data to provide a statistically acceptable classification prediction. Using a pre-processed labelled dataset (comprising 61 event descriptors or dataset features) derived from 475 publicly available aviation investigation reports of prominent safety investigation agencies, a selection of five supervised learning models (e.g. Support Vector Machine, Logistic Regression, Random Forest Classifier, XGBoost and K-Nearest Neighbors) were trained. Each model was evaluated in its ability to classify aviation occurrences into the binary categories of ‘incident’ or ‘serious incident’. This paper showed the best performing ML algorithm was the Random Forest Classifier with accuracy = 0.77, F1 Score = 0.78 and MCC = 0.51 (average of 100 sample runs for each ML model). The study had also explored the effect of applying Synthetic Minority Over-sampling Technique (SMOTE) to correct the slightly imbalanced dataset, and the overall observation ranged from no significant effect to substantial degradation in performance for some of the models after the SMOTE adjustment. This paper serves to provide a practical solution for addressing the variations in occurrence classification standards amongst safety investigation agencies. \\

\providecommand{\keywords}[1]
{
  \small	
  \textbf{\textit{Keywords---}} ##1
}

\noindent \keywords{Artificial Intelligence, Machine Learning, Supervised Learning, Binary Classification, Matthews Correlation Coefficient, Synthetic Minority Over-sampling Technique, Aviation Safety} 

    \end{abstract}\hspace{10pt}
    } 

\begin{document}

\maketitle 


\section{Introduction}

The application of artificial intelligence (AI) and machine learning (ML) in aviation, particularly in the field of aviation safety, has been gaining prominence in recent years with researchers and engineers discovering new ways to apply AI to solve existing challenges in the aviation industry \autocite{Demir:2024jp}. A popular direction for research in this field has been the focus on the transformer architecture, e.g. generative pre-trained transformer (GPT) and large language models (LLM). Some studies have explored harnessing the GPT’s multi-head attention mechanism through finetuning or prompting techniques for investigation report semantic classification \autocite{New:2024jp,Saunders:2024jp,Chandra:2024jp}. This paper’s implementation, which had pre-dated the availability of and the popular use of modern LLM products like OpenAI’s ChatGPT, maintains that traditional supervised ML models still provide present day efficacy and ease of application in classifying aviation safety occurrences.

\begin{figure*} 
	\caption{Screenshot of the ML web application}
	\includegraphics[width=\linewidth]{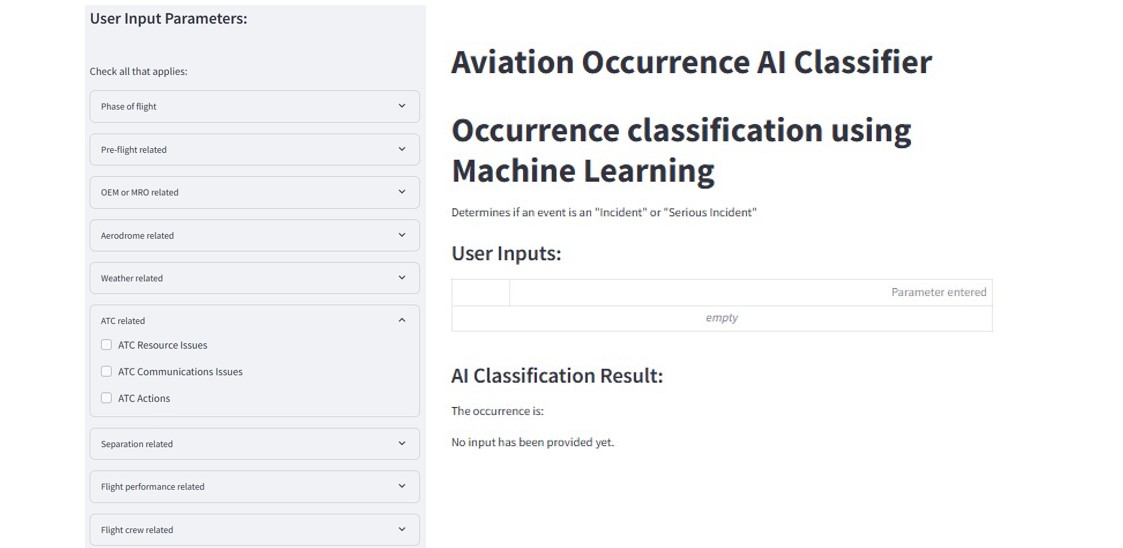}
	\label{fig:webapp}
\end{figure*}

Investigations of aviation safety occurrences are governed by the International Civil Aviation Organization (ICAO) and the guidance material of its Annexes (a set of 19 documents that form part of the Convention on International Civil Aviation or ‘Chicago Convention’ and this guidance sets international standards and recommended practices for international civil aviation). ICAO’s Annex 13 specifically covers aircraft accident and incident investigations, and it provides the definitions of ‘Accident’ and ‘Serious Incident’ which are two classification types that require a country’s designated safety investigation agency (SIA) to conduct a safety investigation \autocite{ICAO:2024qr}. The ICAO Annex 13’s definition for ‘Accident’ is clearly defined and easy to interpret, so that occurrences that are classified as 'Accident' tend to have very clear reasons for being classified as such. The common difficulty faced by aviation safety investigators concerns the ICAO's Annex 13 definition for ‘Serious Incident’, as its definition covers a broad spectrum of scenarios which can be subjective and open to individual interpretations. This lack of clarity in definitions has given rise to recurring situations where different individuals, even within the same SIA, may classify the same aviation occurrence differently (for example, some individuals may propose classifying an occurrence as an incident while others claiming it to be a serious incident). 

An example of why this difference in opinion during classification is a challenge for SIAs can be seen where if an event is classified as an ‘Incident’, the SIA has no obligation to conduct an investigation into the occurrence, whereas if the event is classified as a ‘Serious Incident’, a mandatory investigation is expected. Hence, differing classification standards and a lack of classification consistency within the SIA or amongst global safety investigation agencies are the chief cause of the non-conduct of a safety investigation. As valuable safety lessons can be learnt from a safety investigation, the non-conduct of an investigation would mean that these lessons would be ‘lost’, particularly for a mis-classified occurrence that would not have the chance of an investigation. So there is a need to reduce decision variation during the "Incident/Serious Incident" classification process. Correct classification of occurrences would also help aviation safety investigators better manage their investigation resources to prioritize urgent occurrences that truly require an investigation, thus fulfilling ICAO’s Annex 13 mandate of conducting a safety investigation with the sole objective of preventing future occurrences.

This paper describes a novel approach to implementing a ML web application that uses Supervised ML algorithms (ML models) to help with the task of determining and classifying an event into either an ‘Incident’ (i.e. no obligation to investigate unless the SIA specifically decides to) or a ‘Serious Incident’ (i.e. to be investigated in accordance to ICAO’s Annex 13). See Figure \ref{fig:webapp}. 

The ML web application will guide aviation safety investigators to accurately classify an initial occurrence by removing decision inconsistencies caused by the susceptibility of human opinion to biasness and error. In determining the best approach to implement and design the ML web application, several challenges need to be addressed, such as the philosophy of how aviation occurrence classification should be conducted, the dataset design and its feature selection, harmonizing the representation of information sourced from different report structures of SIAs and finding the best performing ML model to be used.


\section{Methodologies}

\subsection{Setting Up the Classification Framework}

It is common practice for SIAs to publicly publish their final aviation investigation reports, these reports usually come with an official classification by the SIA and would also feature a wide range of essential information that could be derived for use as labelled data for training our ML models. Some previous attempts to utilize the narrative information from safety investigation reports to train AI and LLMs have shown varying degrees of success \autocite{New:2024jp,Saunders:2024jp,Chandra:2024jp,Zhang:2020jp}. This paper employed a quantitative approach where the models would be trained with labelled data distilled from key contextual information found in safety investigation reports. Using this contextual approach to train our ML models would allow adaptability in handling the various styles of contextual data typically encountered in safety investigation reports. The author built a supervised learning framework that allowed drop-in ML models to be deployed as required, this modularity would facilitate performance comparisons between various ML models and allow the incorporation of the best performing model for our use case. The next challenge is to establish the categorical structure of the expected dataset and the various input parameters that the users of our web application can interact with.

\subsection{Data Classes and Feature Selection}

In order to determine the relevant input parameters that will define our dataset, the author reviewed a wide selection of publicly available aviation investigation reports from prominent accident investigation agencies (for instance, the US National Transportation Safety Board, the UK Air Accidents Investigation Branch, the Australian Transport Safety Bureau and France’s Bureau d'enquêtes et d'analyses pour la sécurité de l'aviation civile) to determine the typically encountered range of possible data classes and their associated features. This initial base study served to scope the final dataset classes and features that will be used to train our ML models, and these would form the final input parameters to be incorporated into our ML web application. In total, 17 data classes and 61 associated features have been identified to be used. See Table \ref{tab:dataclass}.

\begin{table}[h!] 
	\caption{Feature selection}
    \tiny
	\centering

	\begin{tabular}{l l}

		\toprule

		{Data Classes}  & {Associated Features}  \\
		\midrule
		Phase of flight & {•	Pushback Phase}  \\
		&  {•	Taxiing or Towing Phase} \\ 
        &  {•	Take-off or Take-off Go-around (TOGA)}  \\ 
        &  {•	Inflight Phase}  \\
        &  {•	Landing Phase}  \\
        &  {•	Parking or Parked Phase}  \\ \\
		Pre-flight related &  {•	Maintenance related during Groundhandling}  \\
        &  {•	Other Groundhandling (e.g. non-maintenance)} \\ 
        &  {•	Issues with Aircraft Load or its Planning}  \\ \\
        Original Equipment Manufacturer (OEM) or  & {•	OEM/MRO known issues}  \\ 
        Maintenance, Repair and Overhaul (MRO) & \\ \\
	  Aerodrome related & {•	Runway under Special Operations}  \\
		&  {•	Involving taxiways} \\ 
        &  {•	Incursion} \\ 
        &  {•	Excursion} \\ 
        &  {•	Runway Overrun} \\ 
        &  {•	Aerodrome related matters} \\ \\
        Weather related & {•	Turbulence by weather}  \\
		&  {•	Weather} \\ \\
        Air traffic control (ATC) related & {•	ATC resource issues }  \\
		&  {•	ATC communications issues} \\ 
        &  {•	ATC actions} \\ \\
        Separation related & {•	  Loss of separation }  \\
		&  {•	Traffic alert and collision avoidance system} \\ 
        &  {	(TCAS) Resolution Advisory} \\ 
        &  {•	TCAS Traffic Advisory} \\
        &  {•	Enhanced Ground Proximity Warning System} \\ \\
        Flight performance related & {•	  Take-off Performance}  \\
		&  {•	Flight Performance } \\ 
        &  {•	Loss of control (of aircraft) during flight} \\
        &  {•	Minimal Safe Altitude breached } \\
        &  {•	Landing configuration or performance issues} \\ \\
        Flight crew related & {•	Flight crew resource issues}  \\
		&  {•	Flight crew communications issues} \\ 
        &  {•	Flight crew actions} \\ \\
        Input error or Omission  & {•	Incorrect input was made by any party }  \\ \\
        Aircraft damage assessment  & {•	Aircraft Damage Replaceable  }  \\
		&  {•	Aircraft Damage Minor Repair } \\ 
        &  {•	TailStrike } \\ 
        &  {•	Foreign object damage (FOD) } \\ 
        &  {•	Birdstrike } \\
        &  {•	Collision } \\
        &  {•	Near collision } \\ \\
        Aircraft system related  & {•	Landing Gears  }  \\
		&  {•	Hydraulic System } \\ 
        &  {•	Fuel System } \\ 
        &  {•	Electrical System } \\ 
        &  {•	Flight Control System	 } \\
        &  {•	Electronic or Avionics related } \\ \\
        Engine issues  & {•	  Engine failure or not usable }  \\
		&  {•	Other engine Issues  } \\ 
        &  {•	Engine damage } \\ \\
        Parts liberated from aircraft/engine  & {•	Any part came off the aircraft or engine  }  \\ \\
        Fire/Smoke/Odour related  & {•	Smoke Fumes Odour related }  \\
		&  {•	Fire Indication Alerts } \\ 
        &  {•	Fuel System } \\ 
        &  {•	Fire indication persist despite actions  } \\ 
        &  {•	Engine fire } \\
        &  {•	Other types of fire } \\ \\
        Pressurisation related issues  & {•	  Pressurisation issue}  \\
		&  {•	Emergency oxygen use  } \\ \\
        Incapacitation/Injuries  & {•	 Incapacitation}  \\
        &  {•	Injuries} \\ \\
		\bottomrule
	\end{tabular}
     
	\label{tab:dataclass}
\end{table}

\subsection{Data Sampling}

The dataset used to train and validate the ML models were derived from 475 aviation safety events of varying nature that had been investigated by various SIAs from 2011 to 2022 (only reports classified as incidents or serious incidents were used). Depending on the nature of the safety event and if the event had been classified as an ‘incident’ or a ‘serious incident’, all pertinent facts and contextual information found to be present in each event would be recorded according to the previously determined list of data classes and features. 

For example, if an investigation report describes an event as a serious incident due to runway excursion during landing in poor weather, we would input the following into our database: \\

        (where '1' is affirmative for the feature)
\begin{itemize}
	\item ‘1’ for Phase of flight (Landing Phase), 
	\item ‘1’ for Aerodrome related (Excursion), 
	\item ‘1’ for Weather related (Weather), 
	\item ‘1’ for Flight crew related (various options, as identified in report)
    \item ‘1’ for Aircraft damage assessment (various options, as identified in report)
    \item ‘1’ for ‘Serious Incident’ (this is the output data)
\end{itemize}

\begin{figure*} 
	\caption{Sample prediction on the ML web application}
	\includegraphics[width=\linewidth]{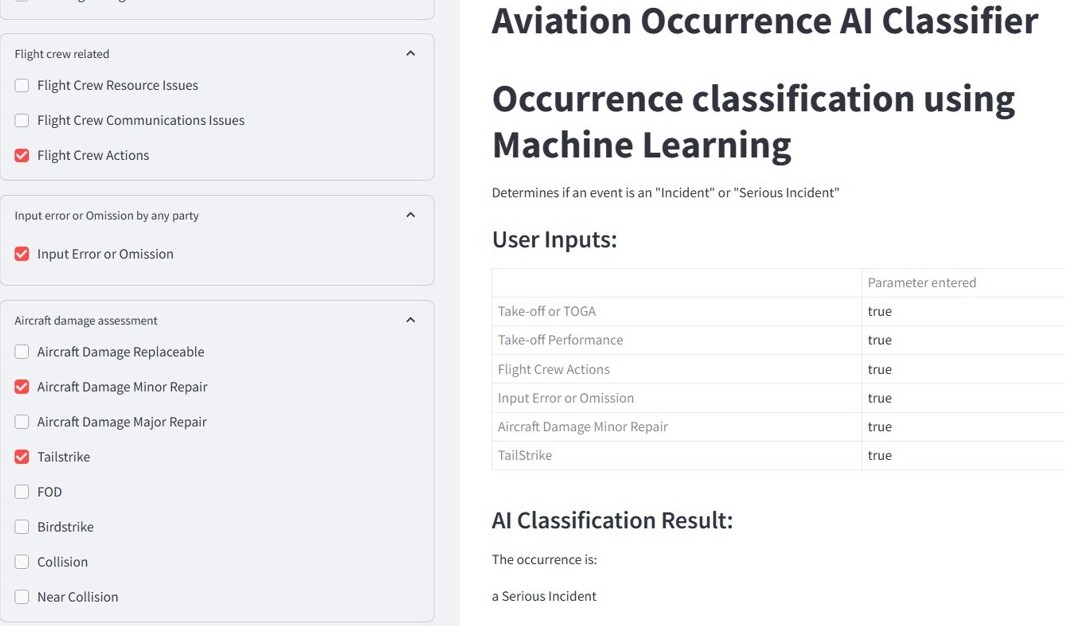}
	\label{fig:pred}
\end{figure*}

\subsection{Training and Tuning the ML Models}

ML models tend to be commonly categorized into supervised, unsupervised or reinforcement learning types. This paper used five types of supervised learning models as test candidates for the classification task, the models used were Support Vector Machine (SVM), Logistic Regression (Log R), Random Forest Classifier (RFC), XGBoost (XGB) and K-Nearest Neighbors (KNN). Models were trained with the dataset split in a 80:20 ratio of training data and testing (validation) data. This means that, for each trial run of the ML model, a randomized 80\% segment of the dataset would be used to train the model so that it can form a correlation between the combinations of input features and the associated output (which in our case is either an ‘incident’ and a ‘serious incident’). The remaining 20\% of the dataset would be used to validate the trained model’s ability to classify outputs correctly. \\ 
Notably, each ML model has its own internal settings (or set of configurable variables) called hyperparameters that control the model’s learning ability and affects its performance. Hyperparameter tuning is an iterative process that allows the identification of the optimal internal setting that produces the maximum performance, this would be accomplished by repeated searching by trial and error. For each trial run of the ML model, this process of hyperparameter tuning would take place to find the optimal setting for each randomized dataset being used, and this optimal setting would be used for model training. When the data validation phase begins, the ML model would have been trained using its optimal hyperparameters so as to ensure best performance. This process is repeated each time a ML model’s performance is measured, a sampling size of 100 measurements was collected for each model. 

\subsection{Applying the Trained Model}

Once we have determined the best performing ML model, we would select that algorithm to be re-trained with 100\% of the dataset and to be applied to the ML web application for deployment. When using the ML web application to make a prediction, the user would select from the drop-down menu the data classes and features that match the pertinent facts and contextual information collected about the occurrence to be classified. The ML web application would assess these inputs and provide its prediction output of 'Incident' or 'Serious Incident'. See Figure \ref{fig:pred}. 

\subsection{Managing the Imbalanced Dataset}

A review of our dataset showed the presence of slight imbalance (60:40 ratio favoring the ‘incident’ category). It is known that imbalanced datasets can lead to biased models that perform poorly for the minority class category and overfits the majority class category. Despite our 60:40 dataset still being regarded as fairly balanced, it is not known if dataset re-balancing could improve model performance for our use case \autocite{Rohan:2020jp}.

In order to balance the dataset, a form of over-sampling adjustment to the minority group called Synthetic Minority Over-sampling Technique (SMOTE) was applied to the data split for training \autocite{Chawla:2002jp}. All models would be trained with this newly adjusted training data (it is to be noted that SMOTE was not applied to the testing data as this will introduce biasness to cause overfitting). The performance of the various ML models being trained with the original dataset and the SMOTE adjusted dataset (using k-nearest neighbors, k=1 and k=5) were compared. This comparative study would provide an understanding if our slightly imbalanced dataset required adjustment to improve model performance.

\subsection{Measuring ML Models’ Performance}

\begin{figure} 
	\caption{A confusion matrix with the four possible outcomes}
	\includegraphics[width=\linewidth]{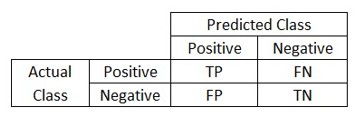}
	\label{fig:confmatrix}
\footnotesize
where:
\begin{itemize}
	\item TP (true positive) is the number of correctly predicted positive cases 
	\item TN (true negative) is the number of correctly predicted negative cases
	\item FP (false positive) is the number of incorrectly predicted positive cases
	\item FN (false negative) is the number of incorrectly predicted negative cases
\end{itemize}
\end{figure}

A common gauge for validating the performance of a classification model is the Confusion Matrix, which is a table that visualizes the model’s correct (true) and incorrect (false) predictions. See Figure \ref{fig:confmatrix}.

To validate and compare the performance between our ML models, we will use various performance metrics derived from the components of the confusion matrix. The three metrics to be used in this study are 1) accuracy, 2) F1 Score and the Matthews Correlation Coefficient (MCC).

\subsection{Accuracy}
Model accuracy in machine learning refers to the proportion of correct predictions (i.e. true positives and true negatives) made by a model out of all predictions made. It's a key metric for general evaluation of model performance. Accuracy ranges from 0 to 1, with 1 being the best accuracy for the model and would indicate good predictive capability. \\

\begin{equation*}
Accuracy = \frac{TP + TN}{TP + FP + TN + FN}
\end{equation*}

\subsection{F1 Score}
F1 Score is a metric that measures the performance for classification models by taking into account the average of two components, precision and recall. 

Precision measures the proportion of true positive predictions among all positive predictions made by the model, granted some positive predictions could be incorrect or false positives. \\

\begin{equation*}
Precision = \frac{TP} {(TP + FP)}
\end{equation*} \\

Recall measures the proportion of true positives predictions among all actual positive instances in the dataset, granted some negative predictions could be incorrect or false negatives. \\

\begin{equation*}
Recall = \frac{TP} {(TP + FN)}
\end{equation*} \\

 F1 Score can be used to evaluate models when there is imbalanced data. In imbalanced datasets, the accuracy score may be misleading if only predicting the majority class as a high accuracy score is still maintained. F1 Score provides a holistic view of model performance by considering both false positives and false negatives. A F1 Score of 1 signifies the best performance and 0 signifies the worst performance.

\begin{equation*}
\begin{split}
F1\; Score & = \frac{2} {[ (1/Precision) + (1/Recall)]} \\ \\
& = \frac{2 \times Precision \times Recall}{Precision + Recall} \\ \\
& = \frac{TP}{TP + 0.5(FP + FN)}
\end{split}
\end{equation*} \\

\begin{table*} [ht]
	\caption{Model comparison (Original Dataset vs Dataset with SMOTE applied)}
\small
\centering
\begin{tabular}{
  l 
  *{9}{S[table-format=1.3]}
  S[table-format=1]
  S[table-format=1.2]
}

\toprule
\multicolumn{1}{c}{\makecell {Model Sample Runs \\ (n=100)}} &
  \multicolumn{3}{c}{Accuracy}  &
  \multicolumn{3}{c}{F1 Score} &
  \multicolumn{3}{c}{MCC} \\
\cmidrule(lr){2-4} \cmidrule(lr){5-7}  \cmidrule(lr){8-10}
  & {Average} & {t} & {\makecell {p-value\\(p<0.005)\textsuperscript*}} & {Average} & {t} & {\makecell {p-value\\(p<0.005)\textsuperscript*}} & {Average} & {t} & {\makecell {p-value\\(p<0.005)\textsuperscript*}}\\
\midrule
RFC (orig)  & 0.7723 &  &  & 0.7807 &  &  & 0.5138 &  &  \\
RFC SMOTE k=1 & 0.7538 & 3.4402 & * & 0.7578 & 4.3565 & *
 & 0.4759 & 3.1761 & *  \\
RFC SMOTE k=5  & 0.7445 & 5.1386 & * & 0.7498 & 5.8805 & * & 0.4550 & 4.8589 & *\\ \\
XGB (orig) & 0.7444 &  &  & 0.7470 &  &  & 0.4578 &  & \\
XGB SMOTE k=1  & 0.7340 & 2.0183 & 0.0449 & 0.7349 & 2.3389 & 0.0203 & 0.4420 & 1.4479 & 0.1492 \\
XGB SMOTE k=5 & 0.7326 & 2.2823 & 0.0235 & 0.7342 & 2.5023 & 0.0131 & 0.4366 & 1.9076 & 0.0578 \\ \\
Log R (orig)  & 0.7399 &  &  & 0.7445 &  &  & 0.4451 &  &  \\
Log R SMOTE k=1  & 0.7397 & 0.0384 & 0.9694 & 0.7388 & 1.0161 & 0.3108 & 0.4623 & -1.4781 & 0.1410 \\
Log R SMOTE k=5  & 0.7382 & 0.3055 & 0.7603 & 0.7376 & 1.2363 & 0.2178 & 0.4590 & -1.2000 & 0.2316 \\ \\
SVM (orig)  & 0.7317 &  &  & 0.7343 &  &  & 0.432 &  &  \\
SVM SMOTE k=1  & 0.7247 & 1.1776 & 0.2404 & 0.7265 & 1.3207 & 0.1881 & 0.4215 & 0.8241 & 0.4109 \\
SVM SMOTE k=5  & 0.7186 & 2.3347 & 0.0206 & 0.7201 & 2.5356 & 0.0120 & 0.4084 & 1.9871 & 0.0483 \\ \\
KNN (orig)  & 0.6940 &  &  & 0.7129 &  &  & 0.3324 &  &  \\
KNN SMOTE k=1  & 0.6667 & 5.0280 & * & 0.6742 & 6.9896 & * & 0.2903 & 3.4411 & * \\
KNN SMOTE k=5  & 0.6655 & 5.0423 & * & 0.6705 & 7.3367 & * & 0.2953 & 2.9079 & * \\
\bottomrule
\end{tabular}
\label{tab:results}
\end{table*}

\subsection{Matthews Correlation Coefficient (MCC)}
The MCC is another performance metric suited to evaluate binary classification models with imbalanced dataset, it achieves this by taking into account all four quadrants of the confusion matrix (true positives, true negatives, false positives, and false negatives). The MCC ranges from -1 to +1, with +1 being a perfect prediction, 0 being no better than random chance and -1 being an inaccurate prediction. It has been argued that the MCC is a more suitable metrics to demonstrate performance sensitivity for binary classification than accuracy, F1 Score or the commonly used receiver operating characteristic curve (ROC) curve \autocite{Chicco:2021jp,Chicco:2023jp}. This paper has chosen to adopt the MCC as one of its metrics to compare ML model performance. \\ \\

${\scriptsize MCC = \frac{(TP \times TN - FP \times FN) }{ \sqrt{(TP + FP) \times (TP + FN) \times (TN + FP) \times (TN + FN)} } }$ \\


\section{Results}

Overall best performing ML model was observed to be the RFC algorithm achieving an average of 0.77 for accuracy, 0.78 for F1 Score and 0.51 for MCC. The lowest scoring performer was the KNN algorithm with the lowest performance in all metric categories. See Table \ref{tab:results}.

The consolidated results table shows the comparison of performance across three performance metrics between ML models that had been trained with the original dataset (with the innate slight imbalance) and the same models trained with synthetically corrected datasets having two types of SMOTE adjustments (i.e. for k=1 and k=5). P-values provide further information if the performance of the SMOTE adjusted model is significantly different (i.e. if p<0.005) from the performance of the original dataset model.

In terms of performance metrics (accuracy, F1 Score and MCC), it is observed that the SMOTE adjustments had significantly decrease performance for RFC and KNN. No significant change to performance metrics can be observed for XGB, Log R and SVM.


\section{Discussion}

\begin{table*} [t!]
	\caption{Performance Benchmarking}

	\centering
 
        \small
 
	\begin{tabular}{l c c c}

		\toprule

		\cmidrule(l){1-4}  
		{Case study} & {\makecell {ML web application \\(machine)}} & {\makecell {Discussion by participants\\(humans)}} & {\makecell {Case as classified by the \\investigation agency (actual)}} \\
		\midrule
		Case 1 & {Incident} & {Incident} & {Incident} \\ \\
		Case 2 & {Serious Incident} & {Serious Incident} & {\makecell {Event was investigated but \\no information on classification}} \\ \\
            Case 3 & {Incident} & {\makecell {Divided \\(majority voted as Incident)}} & {\makecell {Event was investigated but \\no information on classification}} \\ \\
            Case 4 & {Incident*} & {Serious Incident} & {\makecell {Event was investigated but \\no information on classification}}\\ \\
            Case 5 & {Serious Incident} & {Serious Incident} & {\makecell {Initially classified as Incident, \\reclassified as Serious Incident due \\to more information being available}} \\ \\
            Case 6 & {Incident*} & {Serious Incident} & {\makecell {Initially classified as Incident, \\reclassified as Serious Incident for\\ safety awareness reasons}} \\ \\
            Case 7 & {Incident*} &  {\makecell {Divided \\(majority voted as Serious Incident)}} & {Incident} \\ \\
            Case 8 & {Incident} & {Incident} & {No information available} \\ \\

		\bottomrule
	\end{tabular}
    *Mismatch with either participants’ discussion or investigation agency
 
	\label{tab:ecactrial}
\end{table*}

As observed from Table \ref{tab:results}, the best performing ML model was the RFC algorithm, followed by XGB, Log R, SVM and lastly KNN. Basing on accuracy and F1 Score, the RFC showed a strong performance (>75\% for both metrics), and we can corroborate this performance with its strong MCC result (over 0.5 in a -1 to +1 range, +1 being best performance) which is regarded as a dependable indicator of predictive capability \autocite{Chicco:2021jp}. The RFC is an algorithm that uses multiple decision trees applied on data to generate many prediction outputs, the most voted output would be the final prediction \autocite{Jain:2024jp}. The RFC model can be computationally intensive for large datasets as more decision trees are created, however, this issue was not encountered for our smaller dataset and data processing time did not differ much between the ML models (except for XGB). This is beneficial for our ML web application as a short compute time would provide a faster prediction response to the user (as compared to XGB which was the most time and resource consuming model to run). Studies have shown that RFC is better at handling high dimensional data and is more resilient to overfitting than KNN \autocite{Diogo:2021jp}. In view of the high dimensional dataset used in this study, our results showing better performance of RFC over KNN could imply a better handling of the high dimensional dataset.

Initial concerns that the slightly imbalanced dataset might result in biasness and have reduced predictive performance in the models were unfounded. It was seen that the application of synthetic data augmentation (SMOTE) did not provide significant improvement. In fact, the comparison of model performance trained with original dataset and the SMOTE adjusted datasets had showed that SMOTE had significantly reduced the average performance for RFC and KNN, and had no significant effect on XGB, Log R and SVM. One possible explanation of why the SMOTE adjustment did not improve the performance of any of the ML algorithms could be due to its favorable performance with low dimensional data \autocite{Blagus:2013jp}. As such, the high dimensional dataset used in our study could have caused SMOTE adjustment to introduce biasness and noise to the minority class synthetic data points. The interpretation of our results demonstrated that SMOTE did not confer any benefit in performance even after reducing class imbalance, and might even have introduced some degree of biasness. As a result of this finding, it was actually more beneficial to use the original dataset in the final classifier evaluation.

\subsection{Performance Benchmarking (ML web application vs Human predictions)}

This paper’s ML web application implementation was presented at the European Civil Aviation Conference Air Accident and Incident Investigation Group of Experts (ECAC ACC) workshop in Bratislava, Slovakia on 25\textsuperscript{th} April 2023 \autocite{Bryan:2023jp}. The conference conducted a test trial using actual aviation case studies where the predictions of the ML web application was compared with the opinions of the conference participants (sample size of 75 participants, divided into 9 groups) \autocite{ECAC:2023jp}. 

Firstly, the ML web application was tasked to classify the eight case studies provided by the conference organizers and its predictions were recorded. The case studies used were actual investigations volunteered by sponsor investigation agencies. To promote an unbiased trial, the actual classification information for the case studies were initially withheld. Secondly, the conference participants (many of whom can be regarded as industry leaders or subject matter experts in the field of accident investigation) were provided the same eight case studies to discuss in their groups and to deliberate on the most suitable classification for each case. Lastly, the prediction result from the ML web application and the discussion results by the participants were compared with the actual case classification by the investigation agencies. The consolidated results are shown in Table \ref{tab:ecactrial}.

Overall, the predictions of the ML web application were shown to model quite similarly to the discussion outcome of the participant groups with only one prediction being divergent (i.e. Case 4) and two predictions being divided between participants and the investigation agency (i.e. Case 6 and Case 7). For Case 4, while the all participants agreed that the occurrence should be classified as serious, there were several reasons provided by the participants to justify their classification decision, such as flight crew non-compliance, an interest to find out more facts through an investigation, the technicality of aircraft weight that required an investigation etc. This highlighted the variability of the human decision process and that even a unified decision can arise from varied human opinions. For Case 6, the ML web application’s ‘incident’ prediction aligned with investigation agency’s initial ‘incident’ classification based on initial evidence, subsequent upgrade to ‘serious incident’ was due to the decision to promote awareness of the safety issue through the conduct of a public investigation. Case 6 highlighted the flexibility that can be exercised by investigators to make a conscientious decision for conducting an investigation regardless of how the case was classified. For Case 7, the difference between the participants’ discussion outcome and the investigation agency’s classification was attributed to the differing interpretations of existing safety standards, and this understanding varied from person to person. Particularly for Case 7, where differing understanding gave rise to varying classification results, the use of the ML web application would have been useful in providing a baseline opinion.

\section{Conclusions}

This paper demonstrated the feasibility of using the proposed supervised learning framework to solve real-world aviation safety challenges such as the classification of aviation occurrences. The use of ML models trained on an established database of investigation reports helps remove inherent biasness and inconsistency caused by human opinion, this is crucial when safety investigators are deciding whether to institute an investigation or not. This paper’s implementation in the form of a web application has been shown to be practical and easy to be deployed.

When adjusting the slightly imbalanced dataset with SMOTE adjustment, it was observed that this form of synthetic data augmentation to the dataset (for both k=1 and k=5) did not confer any significant benefit to model performance when compared to that of the original dataset. It was considered that SMOTE did not perform well when used on a high dimensional dataset. As such, the original unadjusted dataset was used for the ML model evaluation. 

The paper revealed that the Random Forest Classifier produced the best results over the three performance metrics (accuracy = 0.77, F1 Score = 0.78, MCC = 0.51). 

Considering the potential of our ML web application as demonstrated by actual performance benchmarking and its ability to assist safety investigators to classify aviation occurrences, there is scope to further improve it and incorporate other facets of AI. It is hoped that the practical approach presented in this paper can serve as a useful resource to guide further innovations in aviation safety.


\subsection{Abbreviations}

\small
\begin{tabular}{ll}
AI &   Artificial Intelligence\\
FN &   False negative\\
FP &   False positive\\
GPT &  Generative pre-trained transformer\\
ICAO & International Civil Aviation Organization\\
KNN &  K-Nearest Neighbors\\
LLM &   Large language model\\
Log R &   Logistic Regression\\
MCC &   Matthews Correlation Coefficient\\
ML &   Machine learning\\
RFC &  Random Forest Classifier\\
SIA & Safety investigation agency\\
SMOTE & Synthetic Minority Over-sampling Technique\\
SVM &   Support Vector Machine\\
TN &   True negative\\
TP &   True positive\\
XGB &   XGBoost\\

 \end{tabular}


\subsection{Author Contributions}

The author contributed to the study conception and design, software programming and development, material preparation, data collection and analysis. Draft of the manuscript, its subsequent revisions and final approval were by the author.


\subsection{Declaration}

The author declares no conflict of interest.


\subsection{Acknowledgments}

The author would like to thank Steven G. K. Teo for his support and Davide Chicco for his insightful scientific suggestions.


\printbibliography 


\end{document}